\documentclass[11pt]{article}
\usepackage[margin=1in]{geometry}
\usepackage{amsmath,amssymb,amsthm}
\usepackage{booktabs}
\usepackage{graphicx}
\usepackage{setspace}
\usepackage{enumitem}
\usepackage{xcolor}
\usepackage[round,authoryear]{natbib}
\usepackage[colorlinks=true,linkcolor=blue,citecolor=blue,urlcolor=blue]{hyperref}
\usepackage{xurl}                 

\newtheorem{proposition}{Proposition}
\newtheorem{theorem}{Theorem}
\newtheorem{corollary}{Corollary}
\theoremstyle{definition}
\newtheorem{definition}{Definition}
\newtheorem{assumption}{Assumption}
\theoremstyle{remark}
\newtheorem{remark}{Remark}

\newcommand{\E}{\mathbb{E}}
\newcommand{\Prob}{\mathbb{P}}
\newcommand{\indep}{\perp\!\!\!\perp}
\newcommand{\IIE}{\mathrm{IIE}}
\newcommand{\IDE}{\mathrm{IDE}}

\title{When Reconstruction Is Not Enough: Doubly Robust Counterfactual and Interventional Mediation Analysis with Learned Multimodal Representations}

\author{
Cong Cao\\
Department of Biostatistics, Yale School of Public Health, New Haven, CT 06520, USA\\
Department of Surgery,
Stanford University School of Medicine, Stanford, CA 94305, USA\\
Corresponding author: Cong Cao\\
Email: \texttt{congc@stanford.edu}
}
\date{}

\begin{document}

\maketitle

\begin{abstract}
Learned representations of high-dimensional multimodal health records are increasingly used to adjust for confounding, yet they are usually trained to reconstruct the data rather than to preserve the information that causal identification requires. We study two families of estimands: counterfactual outcomes under alternative treatments, including phenotype-specific effects and patient-level counterfactual risk profiles of the kind digital twins provide, and interventional direct and indirect effects under exposure-induced intermediate confounding. We give conditions under which a representation preserves identification and show that interventional mediation is more demanding than counterfactual comparison. We prove that no training objective that depends only on the covariates, whether reconstruction, contrastive, or physics-informed, can certify that a representation is sufficient, and in a linear structural model we show that a representation can lose an arbitrarily small share of reconstruction accuracy yet induce arbitrarily large bias. We decompose the error of cross-fitted doubly robust estimators into a first-order representation bias, which neither double robustness nor cross-fitting removes, and a second-order remainder of nuisance-estimation errors, so coverage deteriorates as the sample size grows when the representation is insufficient. We propose a workflow combining modality-specific variational autoencoders trained outside the estimation sample, cross-fitted doubly robust and one-step estimators, locked phenotypes, select-then-confirm pathway analysis, portable representations for external replication, and diagnostics for representation insufficiency, exposure leakage, and sensitivity. In simulations, a representation that discarded a rare confounder gave interventional indirect-effect intervals with 2\% coverage at $n=3,000$ when the confounder was moderately strong, reconstruction loss did not identify the problem, and a simple predictive diagnostic flagged it in 100\% of replicates.
\end{abstract}

\noindent\textbf{Keywords:} causal inference; interventional effects; intermediate confounding; mediation analysis; representation learning; variational autoencoder; insufficient representations; counterfactual prediction; double machine learning; physics-informed learning; digital twins

\section{Introduction}

Electronic health records, biobanks, and participant surveys now describe patients through many data types at once: diagnoses, medications, laboratory values, vital signs, social and behavioral measures, and genomes. Causal questions asked of these data, such as whether a treatment works better for some patients than for others, or how much of an exposure's effect on a later outcome runs through an early intermediate change, require adjustment for confounding by many of these characteristics simultaneously. A common strategy is to compress the multimodal record into a low-dimensional representation, for example with modality-specific variational autoencoders (VAEs) \citep{kingma2014}, and to use that representation as the adjustment set in downstream causal analyses \citep{cao2026multimodal,cao2026cognitive}. The appeal is practical: nuisance models with thousands of sparse, correlated, and partly missing features are hard to fit, whereas a handful of learned dimensions is easy to work with.

The difficulty is that a VAE is trained to reconstruct its inputs. Nothing in that objective requires the representation to retain the information that identifies a causal effect, namely information about how the exposure was assigned, how intermediate variables respond, and how the outcome is generated. A variable can matter greatly for confounding while contributing little to reconstruction, for instance a rare diagnosis among hundreds of codes. More generally, good predictive performance does not by itself yield valid causal estimates \citep{westreich2020,naimi2023}. In related settings we found that learned components that predict or reconstruct well can still yield poor causal answers: low nuisance-function prediction error did not identify the better causal estimator \citep{cao2026prediction}; learned exposure mappings with near-identical predictive accuracy implied different spillover effects \citep{cao2026interference}; and different representation-learning objectives recovered different latent structures from the same data \citep{cao2026objectives}. For conditional average treatment effects, \citet{melnychuk2024} showed that low-dimensional representations can induce confounding bias and derived bounds for it.

Mediation analysis is more demanding still. When an exposure affects a confounder of the mediator--outcome relation, natural direct and indirect effects are not identified, and interventional effects are used instead \citep{vanderweele2014,vansteelandt2017}. Their identifying functional involves the exposure mechanism, the distribution of the intermediate confounder, the mediator distribution, and the outcome regression, and efficient estimators combine all of them \citep{diaz2021,diaz2024correction,benkeser2021}. A representation must preserve the information needed by every one of these nuisance functions. Interventional effects also depend on the covariates used to define the random draw of the mediator, so an insufficient representation changes the target itself, not only the estimate.

Representation learning for treatment effects is well developed \citep{shalit2017,louizos2017,shi2019}, and recent work characterizes which information in a learned representation suffices for efficient adjustment \citep{christgau2024}. Deep-learning approaches to mediation include debiased semiparametric estimators \citep{xu2022deepmed}, VAE-based models with hidden confounders \citep{cheng2022cmavae}, VAEs for multi-dimensional and indirectly observed mediators \citep{jiang2023imavae}, and joint representations of covariates and mediators within cross-fitted estimators of natural effects \citep{bo2026mediencoder}. These approaches either target natural effects or assume that the learned representation is sufficient. To our knowledge, the consequences of representation insufficiency for interventional effects under intermediate confounding, and practical diagnostics for it, have not been studied.

This paper makes six contributions. First, we give sufficiency conditions under which a learned representation preserves identification of counterfactual means and of interventional effects, and we show that the interventional conditions are strictly more demanding (Section~\ref{sec:sufficiency}). Second, we prove that no training objective that depends only on the covariates, including reconstruction, contrastive, and physics-informed objectives, can certify sufficiency (Section~\ref{sec:sufficiency}), and, for a linear structural model, we derive the bias of interventional indirect effects induced by an insufficient representation and show that no function of reconstruction error bounds it (Section~\ref{sec:linear}). Third, we decompose the error of cross-fitted doubly robust estimators that use a learned representation into a first-order representation bias and a second-order remainder, and we show that coverage deteriorates with sample size when the representation is insufficient (Section~\ref{sec:decomposition}). Fourth, we describe a workflow that applies one representation to counterfactual treatment comparisons, locked phenotypes, interventional mediation, select-then-confirm pathway analysis, and portable external replication, including time-indexed multimodal records with a pre-exposure window (Section~\ref{sec:workflow}). Fifth, we propose diagnostics for representation insufficiency, exposure leakage, and representation sensitivity, together with a rule for choosing the representation by nuisance loss rather than reconstruction loss (Section~\ref{sec:diagnostics}). Finally, we evaluate these methods in a simulation study with a tunable information-loss mechanism (Sections~\ref{sec:simulation} and~\ref{sec:results}).

\section{Setup and estimands}
\label{sec:setup}

\subsection{Data and potential outcomes}

We observe $n$ independent copies of $O=(W,A,Z,M,Y)$. Here $W=(W^{(1)},\ldots,W^{(K)})$ collects pretreatment covariates from $K$ modalities, for example diagnoses, medications, laboratory values, vital signs, and social and behavioral measures; $A\in\{0,1\}$ is a treatment or exposure; $Z\in\{0,1\}$ is an intermediate confounder that may be affected by $A$; $M$ is a mediator measured after $A$, such as early weight change; and $Y$ is an outcome measured after $M$. For counterfactual treatment comparisons we also allow a vector of binary failure indicators $Y^{(1)},\ldots,Y^{(F)}$ in place of $Y$. Let $Y(a)$ denote the potential outcome under $A=a$, $M(a)$ the potential mediator, and $Y(a,m)$ the potential outcome under $A=a$ and $M=m$.

For a generic adjustment set $V$, which will be either $W$ or a representation $\phi(W)$, define the nuisance functions
\begin{align*}
g(a\mid v)&=\Prob(A=a\mid V=v), & q(z\mid a,v)&=\Prob(Z=z\mid A=a,V=v),\\
p(m\mid a,z,v)&=f_{M\mid A,Z,V}(m\mid a,z,v), & b(a,z,m,v)&=\E(Y\mid A=a,Z=z,M=m,V=v),\\
\mu_a(v)&=\E(Y\mid A=a,V=v), & \bar p(m\mid a,v)&=\textstyle\sum_{z}p(m\mid a,z,v)\,q(z\mid a,v).
\end{align*}
Here $f_{M\mid A,Z,V}$ is the conditional density of $M$, and $\bar p(m\mid a,v)$ is the density of $M$ given $(A,V)=(a,v)$ with $Z$ marginalized.

\subsection{Estimands}

\paragraph{Counterfactual outcomes under alternative treatments.} The first family compares treatment options for the same patients. Its targets are the counterfactual mean $\psi(a)=\E\{Y(a)\}$; the phenotype-specific counterfactual mean $\psi(a;s)=\E\{Y(a)\mid S=s\}$ and effect $\tau(s)=\psi(1;s)-\psi(0;s)$, where $S=s(W)$ is a pretreatment phenotype (Section~\ref{sec:phenotypes}); and the counterfactual risk profile $r^{(f)}_a(w)=\Prob\{Y^{(f)}(a)=1\mid W=w\}$ for failure mode $f$ under treatment $a$. The risk profile answers the question a clinician asks at the first visit: for this patient, which failure is most likely under each option?

\paragraph{Interventional direct and indirect effects.} The second family decomposes the effect of $A$ on $Y$ through $M$. Following \citet{vanderweele2014}, let $G_{a^*\mid W}$ denote a random draw from the distribution of $M(a^*)$ given $W$, and define
\[
\theta_W(a,a^*)=\E\{Y(a,G_{a^*\mid W})\}.
\]
The interventional indirect and direct effects are
\[
\IIE=\theta_W(1,1)-\theta_W(1,0),\qquad \IDE=\theta_W(1,0)-\theta_W(0,0),
\]
which sum to the overall interventional effect $\theta_W(1,1)-\theta_W(0,0)$. In words, the $\IIE$ is the change in the average outcome when only the distribution of the mediator is shifted from its level under $A=0$ to its level under $A=1$, holding the exposure at $A=1$. The subscript $W$ matters: the estimand is defined relative to the covariates used to define the random draw.

\subsection{Identification given the full covariate set}

\begin{assumption}[Identification given $W$]\label{ass:id}
(i) Consistency: $M=M(A)$ and $Y=Y(A,M)$. (ii) $Y(a,m)\indep A\mid W$ and $M(a^*)\indep A\mid W$ for all $a,a^*,m$. (iii) $Y(a,m)\indep M\mid (A,Z,W)$ for all $a,m$. (iv) Positivity: $g(a\mid W)>0$ and $p(m\mid a,z,W)>0$ wherever $p(m\mid a^*,z',W)>0$, almost surely.
\end{assumption}

Under Assumption~\ref{ass:id}, the counterfactual mean is identified by $\psi_W(a)=\E\{\mu_a(W)\}$ and the interventional mean by \citep{vanderweele2014,diaz2021}
\begin{equation}\label{eq:theta}
\theta_W(a,a^*)=\E\Big\{\sum_{z}\int b(a,z,m,W)\,q(z\mid a,W)\,\bar p(m\mid a^*,W)\,dm\Big\}.
\end{equation}
For a representation $\phi$, write $\psi_\phi(a)$ and $\theta_\phi(a,a^*)$ for the same functionals with $W$ replaced by $\phi(W)$ in every nuisance function and in the outer expectation. These are the quantities that a doubly robust estimator using $\phi(W)$ as covariates targets. The question is when they equal $\psi_W(a)$ and $\theta_W(a,a^*)$.

\subsection{Multimodal representations}
\label{sec:representation}

For each modality $k$, a VAE with encoder $q_k(u_k\mid w^{(k)})$ and decoder $p_k(w^{(k)}\mid u_k)$ is trained by maximizing the evidence lower bound
\[
\mathcal L_k=\E_{q_k}\{\log p_k(W^{(k)}\mid U_k)\}-\beta\,\mathrm{KL}\{q_k(U_k\mid W^{(k)})\,\|\,\mathcal N(0,I_{d_k})\},
\]
with likelihoods matched to the data type (Gaussian for continuous laboratory values, Bernoulli for diagnosis indicators, categorical for survey items). The patient representation is the concatenation of posterior means,
\[
\phi(W)=\big(\E_{q_1}(U_1\mid W^{(1)}),\ldots,\E_{q_K}(U_K\mid W^{(K)})\big)\in\mathbb R^{d},\qquad d=\textstyle\sum_k d_k,
\]
which compresses each modality separately before integration, as in our earlier latent-space analyses \citep{cao2026multimodal}. We also consider the augmented representation $\phi^+(W)=(\phi(W),W_{\mathrm{key}})$, in which a short list of prespecified clinical confounders $W_{\mathrm{key}}$ is added unchanged.

Because the representation is itself estimated, where it is trained matters. We distinguish three regimes. Under external training (R1), the encoders are trained on covariate data from individuals outside the analytic sample, for example other participants of the same biobank, and then held fixed. Under fold-specific training (R2), within $K$-fold cross-fitting, the encoders for fold $j$ are trained on the covariates of the other folds, together with the nuisance models for fold $j$. Under full-sample unsupervised training (R3), the encoders are trained on the covariates $W_1,\ldots,W_n$ of the whole analytic sample, without using $A$, $Z$, $M$, or $Y$, and the nuisance models are then cross-fitted.
\paragraph{Time-indexed records.} In many settings the multimodal record is a time series, for example continuous physiological monitoring, device and activity logs, or audio and video recordings of a clinical episode, and the exposure is defined at or after a time zero $t_0$, as in target-trial emulation \citep{hernan2016}. The framework applies with $W$ replaced by the pre-exposure window $W_{<t_0}$: the encoders are trained, and the representation is computed, only from data recorded before $t_0$. Data recorded after $t_0$ may reflect the exposure or its consequences. Including them in the representation conditions on post-exposure variables, can encode the exposure itself (diagnostic D2 in Section~\ref{sec:diagnostics}), and changes the estimand. Post-exposure variables that are affected by the exposure and confound the mediator--outcome relation enter the analysis as intermediate confounders $Z$, not through the representation. When an exposure accumulates over a window, for example as a count or a rate, the window and $t_0$ should be fixed in advance, so that longer or more complex episodes do not receive larger exposure values merely because they last longer.

\paragraph{Mechanism-informed objectives.} The VAE objective can be augmented with penalties that encode mechanistic knowledge, as in physics-informed neural networks \citep{raissi2019}, in which a learned model is also required to satisfy a differential equation; we have used this approach to forecast urban air pollution from traffic and weather \citep{cao2025pbml}. Two kinds of mechanism should be distinguished. A mechanism that describes only the covariates, for example the dynamics of pretreatment physiological trajectories, yields an objective that still depends on the law of $W$ alone and is therefore subject to Proposition~\ref{prop:covonly}: it may make the representation more accurate or more stable, but it cannot by itself make it sufficient. A mechanism that links the exposure, the mediator, and the outcome, for example an energy-balance model of how a change in energy intake under treatment translates into weight change over time \citep{hall2011}, constrains the nuisance functions on which sufficiency depends and may therefore help. Because such an objective uses $A$, $M$, or $Y$, it falls under the rule below, and its benefit should be checked with diagnostic D1 rather than assumed.

Representations whose training loss involves $A$, $Z$, $M$, or $Y$ (supervised or semi-supervised objectives) use outcome information and are admissible only under regimes (R1) or (R2).

\section{When does a learned representation preserve identification?}
\label{sec:theory}

\subsection{Sufficient representations}
\label{sec:sufficiency}

\begin{definition}[Sufficiency]\label{def:suff}
Let $\phi$ be a measurable map of $W$.
\begin{enumerate}[label=(\alph*),leftmargin=*,itemsep=1pt]
\item $\phi$ is \emph{balancing} if $A\indep W\mid\phi(W)$, and \emph{prognostic} if $\mu_a(W)$ depends on $W$ only through $\phi(W)$ for $a\in\{0,1\}$.
\item $\phi$ is \emph{nuisance-sufficient for mediation} if $b(a,z,m,W)$, $q(z\mid a,W)$, and $\bar p(m\mid a^*,W)$ depend on $W$ only through $\phi(W)$, for all $a,a^*,z,m$.
\item $\phi$ is \emph{process-sufficient} if $(A,Z,M)\indep W\mid\phi(W)$.
\end{enumerate}
\end{definition}

Balancing and prognostic representations generalize the propensity score \citep{rosenbaum1983} and the prognostic score \citep{hansen2008} to learned, multivariate summaries.

\begin{proposition}[Identification with a representation]\label{prop:id}
Suppose Assumption~\ref{ass:id} holds.
\begin{enumerate}[label=(\alph*),leftmargin=*,itemsep=1pt]
\item If $\phi$ is balancing or prognostic, then $\psi_\phi(a)=\psi_W(a)$. If, in addition, the phenotype $S$ is a function of $\phi(W)$, then $\psi_\phi(a;s)=\psi_W(a;s)$.
\item If $\phi$ is nuisance-sufficient for mediation, then $\theta_\phi(a,a^*)=\theta_W(a,a^*)$ for all $a,a^*$.
\item If $\phi$ is process-sufficient, then Assumption~\ref{ass:id} holds with $W$ replaced by $\phi(W)$; the conditional law of $M(a^*)$ given $W$ equals its conditional law given $\phi(W)$; and both $\psi_\phi(a)=\psi_W(a)$ and $\theta_\phi(a,a^*)=\theta_W(a,a^*)$.
\item Balancing does not imply $\theta_\phi=\theta_W$: there are laws under which a balancing $\phi$ identifies every counterfactual mean but not the interventional effects.
\end{enumerate}
\end{proposition}

Proofs are in Appendix~\ref{app:proofs}. Part (d) follows from a simple case: if $A$ is randomized, a constant $\phi$ is balancing and $\psi_\phi(a)=\E(Y\mid A=a)=\psi_W(a)$, yet whenever $W$ confounds the mediator--outcome relation, $\theta_\phi(a,a^*)$ is the unadjusted interventional functional and differs from $\theta_W(a,a^*)$. The practical implication is that a representation adequate for comparing treatments can be inadequate for decomposing their effects.

\begin{remark}[The estimand moves with the representation]
Interventional effects are defined relative to the covariates used to draw the mediator. Under process sufficiency the random draws $G_{a^*\mid W}$ and $G_{a^*\mid\phi(W)}$ have the same distribution, so nothing is lost. Without sufficiency, $\theta_\phi(a,a^*)$ is in general neither $\theta_W(a,a^*)$ nor the causal quantity $\E\{Y(a,G_{a^*\mid\phi(W)})\}$, because exchangeability fails given $\phi(W)$.
\end{remark}

\begin{remark}[Reconstruction targets none of these conditions]
The VAE objective depends only on the marginal law of $W$. All conditions in Definition~\ref{def:suff} concern the conditional law of $(A,Z,M,Y)$ given $W$. Two data-generating laws with the same distribution of $W$ therefore yield the same representation, while one may satisfy sufficiency and the other not. Proposition~\ref{prop:covonly} makes this precise for every objective of this kind, and Section~\ref{sec:linear} makes it quantitative.
\end{remark}

\begin{definition}[Covariate-only objective]\label{def:covonly}
A training objective for a representation is \emph{covariate-only} if, for every candidate $\phi$, its value depends on the data-generating law only through the marginal law of $W$. Reconstruction objectives such as the evidence lower bound, contrastive objectives defined on $W$, and physics-informed penalties that constrain only the dynamics of $W$ are covariate-only.
\end{definition}

\begin{proposition}[No covariate-only objective certifies sufficiency]\label{prop:covonly}
Let $W$ take values in a standard Borel space, and let $\phi$ be a representation such that $W$ is not almost surely a function of $\phi(W)$. Then there are three laws $P_0$, $P_1$, and $P_2$ with the same marginal law of $W$, each satisfying Assumption~\ref{ass:id}, such that $\phi$ is process-sufficient under $P_0$, $\psi_\phi(1)\neq\psi_W(1)$ under $P_1$, and $\IIE_\phi\neq\IIE_W$ under $P_2$. In particular, every covariate-only objective takes the same value at $\phi$ under all three laws, so no such objective, and no rule that selects a representation by its value, can distinguish a sufficient representation from an insufficient one.
\end{proposition}

The only representations that are sufficient under every law are those that retain all of the information in $W$, and they provide no compression. Sufficiency therefore has to come from information about the exposure, mediator, and outcome mechanisms, either through the training objective (Section~\ref{sec:representation}) or through diagnostics such as D1 (Section~\ref{sec:diagnostics}).

\subsection{Insufficiency bias in a linear structural model}
\label{sec:linear}

Let $W=(U,R)$, where $U\in\mathbb R^p$ is retained by the representation, $\phi(W)=U$, and $R$ is a scalar that is lost, with $\mathrm{Var}(R)=\sigma_R^2$ and $R\indep U$. (Any lost component can be taken orthogonal to $U$ after projection.) Consider
\begin{align*}
A&=\gamma^\top U+\gamma_R R+\epsilon_A, \\
M&=\alpha A+\delta^\top U+\delta_R R+\epsilon_M, \\
Y&=\vartheta A+\beta M+\lambda^\top U+\lambda_R R+\epsilon_Y,
\end{align*}
with independent Gaussian errors of variances $\sigma_A^2,\sigma_M^2,\sigma_Y^2$, Gaussian $(U,R)$, and no intermediate confounder. The exposure is continuous here so that the result has closed form; then $\IIE(a,a^*)=\alpha\beta(a-a^*)$ and interventional and natural indirect effects coincide.

\begin{proposition}[Closed-form insufficiency bias]\label{prop:linear}
Let
\[
\kappa=\frac{\gamma_R\sigma_R^2}{\gamma_R^2\sigma_R^2+\sigma_A^2},\qquad v_R=\frac{\sigma_R^2\sigma_A^2}{\gamma_R^2\sigma_R^2+\sigma_A^2}.
\]
Adjusting for $\phi(W)=U$ gives $\IIE_\phi(a,a^*)=\tilde\alpha\tilde\beta(a-a^*)$, where
\[
\tilde\alpha=\alpha+\delta_R\,\kappa,\qquad
\tilde\beta=\beta+\frac{\lambda_R\,\delta_R\,v_R}{\delta_R^2v_R+\sigma_M^2},
\]
so the bias per unit contrast is $\tilde\alpha\tilde\beta-\alpha\beta$. The bias of the total effect is $(\lambda_R+\beta\delta_R)\kappa$.
\end{proposition}

The two factors have distinct sources. The exposure--mediator path is biased when the lost component affects both the exposure and the mediator ($\gamma_R\delta_R\neq0$); the mediator--outcome path is biased when it affects both the mediator and the outcome ($\delta_R\lambda_R\neq0$), even if the exposure is randomized ($\gamma_R=0$, so $\kappa=0$ and $v_R=\sigma_R^2$). The second case is the linear counterpart of Proposition~\ref{prop:id}(d).

\begin{corollary}[Reconstruction error does not bound the bias]\label{cor:recon}
Suppose $R$ is a coordinate of $W$ and reconstruction is measured by squared error. Dropping $R$ increases the minimal reconstruction error by exactly $\sigma_R^2$, a quantity determined by the law of $W$ alone. For every $\varepsilon>0$ and $B>0$ there are coefficients such that $\sigma_R^2\le\varepsilon$ and $|\IIE_\phi-\IIE_W|\ge B$ per unit contrast.
\end{corollary}

A representation can therefore reconstruct the data almost perfectly and still yield arbitrarily large bias. A rare diagnosis that strongly affects both treatment choice and early response is the clinical prototype: it contributes little to reconstruction, so a compressed representation has every incentive to drop it.

\subsection{Error decomposition for cross-fitted estimators}
\label{sec:decomposition}

Let $\Psi$ denote either $\psi(a)$, $\psi(a;s)$, or $\theta(a,a^*)$, and let $\hat\phi$ be a learned representation that is independent of the estimation sample, as under regime (R1); under regime (R2) the argument below applies fold by fold. Write $\eta_{\hat\phi}$ for the true nuisance functions of the adjustment set $\hat\phi(W)$, $\hat\eta$ for their cross-fitted estimates, $D_{\hat\phi}$ for the efficient influence function of the functional $\Psi_{\hat\phi}$ treating $\hat\phi$ as fixed, and $\hat\Psi$ for the cross-fitted augmented inverse-probability-weighted or one-step estimator that uses $\hat\phi(W)$ as covariates.

\begin{theorem}[Error decomposition]\label{thm:decomp}
Suppose that, conditionally on $\hat\phi$, the nuisance estimates are $L_2$-consistent, the exposure-propensity and density-ratio estimates are bounded away from zero, and the usual cross-fitting conditions hold. Then
\[
\hat\Psi-\Psi_W
=\underbrace{\big(\Psi_{\hat\phi}-\Psi_W\big)}_{B(\hat\phi):\ \text{representation bias}}
+\underbrace{(\mathbb P_n-\mathbb P)D_{\hat\phi}}_{\text{first-order, mean zero}}
+\underbrace{R_2(\hat\eta,\eta_{\hat\phi})}_{\text{second-order remainder}}
+o_{\mathbb P}(n^{-1/2}).
\]
For $\psi(a)$, $R_2=\E\big[\{\hat\mu_a(\hat\phi)-\mu_a(\hat\phi)\}\{\hat g(a\mid\hat\phi)-g(a\mid\hat\phi)\}/\hat g(a\mid\hat\phi)\big]$. For $\theta(a,a^*)$, $R_2$ is a sum of products of errors in the exposure, intermediate-confounder, mediator, and outcome nuisances, as given by \citet{diaz2021}.
\end{theorem}

The decomposition separates what the estimation procedure can and cannot fix. Double robustness and cross-fitting make $R_2$ and the empirical-process term negligible under rate conditions \citep{chernozhukov2018,zivich2021}, so that $\hat\Psi$ is asymptotically normal around $\Psi_{\hat\phi}$. They do nothing about $B(\hat\phi)$, which is fixed once the representation is fixed and does not shrink as $n$ grows. The bias can be split further as $B(\hat\phi)=(\Psi_{\hat\phi}-\Psi_{\phi^\circ})+(\Psi_{\phi^\circ}-\Psi_W)$, where $\phi^\circ$ is the population limit of the training procedure: the first term reflects estimation of the representation and is small when the encoders are trained on a large external sample; the second reflects insufficiency of the architecture and objective and is not reduced by more training data.

\begin{corollary}[Coverage deteriorates with sample size]\label{cor:coverage}
If $B(\hat\phi)=b\neq0$ and the variance estimator is consistent for $\sigma^2=\mathrm{Var}(D_{\hat\phi})$, the coverage of the Wald interval for $\Psi_W$ is approximately $\Phi(1.96-\sqrt n\,b/\sigma)-\Phi(-1.96-\sqrt n\,b/\sigma)$, which tends to zero as $n\to\infty$.
\end{corollary}

\begin{remark}[Full-sample unsupervised training]
Under regime (R3), $\hat\phi$ depends on the covariates of the estimation sample, so the conditional argument above does not apply directly. The decomposition continues to hold if the encoder class is Donsker or the training algorithm is sufficiently stable, because $\hat\phi$ uses no information about $(A,Z,M,Y)$. Whether this matters in practice is examined in the simulation.
\end{remark}

\begin{remark}[Connection to prediction error]
With $V=W$, no representation, and the partially linear residualized estimator of a constant effect $\tau$, the second-order term reduces to $\E\{\delta_m(\delta_h-\tau\delta_m)\}/\{\mathrm{Var}(X\mid C)+\mathcal E_m\}$ in the notation of \citet{cao2026prediction}. That paper's finding that marginal prediction error does not track causal bias is thus a statement about the second term of Theorem~\ref{thm:decomp}; the present paper adds the first.
\end{remark}

\section{Estimation workflow}
\label{sec:workflow}

The workflow fits one representation and uses it for every causal estimand. Throughout, the sample is split once into a development part $\mathcal D_{\mathrm{dev}}$, used to train encoders under regime (R1) or (R2), choose the representation, and derive phenotypes, and an estimation part $\mathcal D_{\mathrm{est}}$, used for cross-fitted estimation with everything learned on $\mathcal D_{\mathrm{dev}}$ held fixed. All observations of a patient, and of a cluster when observations are clustered, stay on one side of every split.

\subsection{Counterfactual comparisons of treatments}
\label{sec:counterfactual}

With cross-fitted estimates $\hat\mu_a$ and $\hat g$ computed on $\hat\phi(W)$, define the doubly robust pseudo-outcome
\[
\hat\xi_i=\hat\mu_1(\hat\phi_i)-\hat\mu_0(\hat\phi_i)+\frac{A_i-\hat g(1\mid\hat\phi_i)}{\hat g(1\mid\hat\phi_i)\,\hat g(0\mid\hat\phi_i)}\big\{Y_i-\hat\mu_{A_i}(\hat\phi_i)\big\},\qquad \hat\phi_i=\hat\phi(W_i).
\]
Its sample mean estimates the average effect, its mean within $S=s$ estimates $\tau(s)$ for a locked phenotype, and its regression on $\hat\phi$ in a further split gives the doubly robust learner of the conditional effect function \citep{kennedy2023}. Under Proposition~\ref{prop:id}(a) and Theorem~\ref{thm:decomp}, these estimators are consistent for the $W$-adjusted targets when $\hat\phi$ is balancing or prognostic, which is a weaker requirement than in mediation.

The counterfactual risk profile $\hat r^{(f)}_a(\hat\phi)$ is the cross-fitted outcome model for failure mode $f$ in treatment arm $a$, evaluated for every patient under every option, so that each patient receives a profile of predicted failure risks across treatments and the failure most likely under each option. Because the profile is a plug-in quantity without an augmentation term, we check it against doubly robust targets: within bins of predicted risk, the mean of $\hat r^{(f)}_a$ is compared with the doubly robust estimate of $\Prob\{Y^{(f)}(a)=1\mid\text{bin}\}$, which gives a counterfactual calibration plot for each option. Failure modes are treated as separate binary outcomes; competing failures can be handled with a multinomial outcome model without changing the rest of the workflow.

\begin{remark}[Counterfactual risk profiles as digital twins]
A model that predicts each patient's outcomes under every treatment option is what the clinical literature calls a digital twin \citep{bjornsson2020,nasem2024}. The results above state when such a twin can be trusted. For comparisons between treatments, the representation must be balancing or prognostic (Proposition~\ref{prop:id}(a)); if the twin is also used to explain why outcomes differ, through interventional mediation, it must satisfy the stronger conditions of Proposition~\ref{prop:id}(b) or~(c). The counterfactual calibration check above validates a twin's predictions under each option, and landmark updating (Section~\ref{sec:phenotypes}) corresponds to updating the twin with new data, with the caveat that updated predictions condition on the treatment actually received and are therefore not counterfactual comparisons between options.
\end{remark}

\subsection{Interventional mediation}
\label{sec:mediation}

Interventional effects are estimated with the cross-fitted one-step estimator of \citet{diaz2021}, using $\hat\phi(W)$ or $\hat\phi^+(W)$ in place of $W$; the 2024 correction to that paper applies \citep{diaz2024correction}. We write the mediator density ratio in the efficient influence function through Bayes' rule, so that it requires only the probabilities $\Prob(A\mid M,V)$ and $\Prob(Z\mid A,M,V)$ rather than conditional densities (Appendix~\ref{app:eif}). We implement the estimator directly because regime (R2) uses a different covariate matrix in each fold; for adjustment sets that are common to all folds it can be compared with the \texttt{medoutcon} implementation \citep{hejazi2022medoutcon}, which accommodates a single binary intermediate confounder, and settings with several mediators or intermediate confounders can use the extension of \citet{rudolph2024practical}, implemented in \texttt{HDmediation}. The implementation reproduces the true values exactly when the true nuisance functions are supplied, and remains unbiased when the outcome model or the exposure model is deliberately misspecified. Variances are estimated from the efficient influence function, with cluster-level score sums when observations are clustered. When $Y$ is an incident event, follow-up begins after $M$ is measured so that the mediator precedes the outcome.

\subsection{Phenotypes}
\label{sec:phenotypes}

A phenotype is a pretreatment grouping $S=s\{\hat\phi(W)\}$ derived on $\mathcal D_{\mathrm{dev}}$, for example by linking the representation to observed response and failure modes, and then locked. Each phenotype is described by the features of $W$ that distinguish it, so that it can be recognized without running the model. Locking is what makes phenotype-specific inference valid: because $S$ is a fixed function when $\mathcal D_{\mathrm{est}}$ is analyzed, $\tau(s)$ and the phenotype-specific interventional effects $\theta(a,a^*;s)=\E\{\sum_z\int b\,q\,\bar p\,dm\mid S=s\}$ are ordinary functionals, estimated by within-phenotype averages of the corresponding uncentered influence functions. Deriving and evaluating phenotypes on the same data would make $S$ data-dependent and the resulting intervals too narrow.

The added value of the representation is assessed by comparing phenotype and outcome models built on $\hat\phi$ with models built on conventional clinical characteristics and with models using the same data without a learned representation. Phenotype assignments can be updated at later landmarks, such as six weeks or six months after treatment start, by adding early response to the model without changing the phenotype definitions \citep{vanhouwelingen2007}. Updated assignments condition on the treatment actually received and on early response, so they are predictions for monitoring, not counterfactual comparisons between treatments.

\subsection{Select-then-confirm pathway analysis}
\label{sec:selection}

When many candidate exposure--mediator--outcome pathways are examined, as in latent-space mediation analyses with hundreds of combinations \citep{cao2026multimodal}, ranking and inference on the same data inflate the selected effects. We therefore split $\mathcal D_{\mathrm{est}}$ into a screening part and a confirmation part. On the screening part, each candidate interventional indirect effect is estimated, candidates are ranked by $|\widehat{\IIE}_j|/\widehat{\mathrm{SE}}_j$, and the stability of each candidate is measured by its frequency among the top $K$ across bootstrap resamples. The top-ranked stable candidates are then re-estimated on the confirmation part, with a Holm adjustment over the selected set. Because the selected set depends on the screening data only, the confirmation estimates retain their usual asymptotic distribution conditionally on the selection. We describe this step as internal re-testing, not independent replication.

\subsection{Portable representations and external replication}
\label{sec:portable}

Data sources record different variables. A portable representation $\phi_c$ is trained on the variables $W_c$ that all sources share, with a prespecified substitute for any variable that one source records differently, and is applied unchanged elsewhere. Portability has a checkable and an uncheckable part. The checkable part is whether restricting to $W_c$ already loses information in the development source: the difference $\Delta_{\mathrm{port}}=\hat\Psi(\hat\phi)-\hat\Psi(\hat\phi_c)$, computed on the same estimation sample with a bootstrap interval, should be near zero. The uncheckable part is whether conditional exchangeability between populations holds given $\phi_c$ \citep{pearl2014transportability,dahabreh2020}. External replication then compares prespecified quantities between sources: phenotype prevalence, defining features, calibration and discrimination of each outcome model, and performance across age, sex, race and ethnicity, and other prespecified subgroups.

\section{Diagnostics}
\label{sec:diagnostics}

Theorem~\ref{thm:decomp} shows that the representation bias cannot be controlled by the estimator, so it has to be examined directly. We propose four diagnostics, all computed without access to the outcome of the estimation sample beyond what the nuisance models already use.

\paragraph{D1: Nuisance insufficiency.} For each nuisance function required by the estimand ($g$ and $\mu_a$ for counterfactual comparisons; $g$, $q$, the mediator model, and $b$ for mediation), compare the cross-validated loss of a model using $\hat\phi$ with that of the same learner using $(\hat\phi,W)$. Per-observation held-out loss differences give a mean improvement and a standard error. A clear improvement shows that $\hat\phi$ is not nuisance-sufficient for that function (Definition~\ref{def:suff}). The converse does not hold: no improvement is consistent with sufficiency but does not prove it, and the diagnostic cannot detect unmeasured confounding. When $W$ is too high-dimensional for the learner, a penalized learner on $(\hat\phi,W)$ or on $\hat\phi$ plus the residuals of $W$ given $\hat\phi$ can be used.

\paragraph{D2: Exposure leakage and overlap.} A representation of a rich health record may encode the exposure itself, for example when diagnoses, laboratory values, or medications that define the exposure are among its inputs. We report the cross-validated area under the curve for predicting $A$ from $\hat\phi$, the distribution of $\hat g(1\mid\hat\phi)$, and the share of observations with estimated propensities outside $[0.025,0.975]$. An area near one signals a practical positivity violation; the remedy is to remove exposure-defining variables from the encoder inputs, with trimming as a sensitivity analysis.

\paragraph{D3: Representation sensitivity.} We recompute every estimate across latent dimensions, KL weights, random seeds, and representation objectives (VAE, principal components, and a supervised alternative) and report the range. Different objectives can recover different structures from the same data \citep{cao2026objectives}; conclusions that change across this range are fragile.

\paragraph{D4: Choosing the representation.} Reconstruction loss is reported but not used to choose the representation for causal analysis (Proposition~\ref{prop:covonly} and Corollary~\ref{cor:recon}). Instead, on the development data, we choose the architecture, latent dimension, and KL weight that minimize the sum of standardized cross-validated losses of the nuisance functions that the estimand requires. This targets Definition~\ref{def:suff}(b) directly, although it remains silent about confounders that are not in $W$; standard sensitivity analyses for unmeasured confounding of the exposure--outcome and mediator--outcome relations \citep{vanderweele2015book} are reported alongside.

\section{Simulation study}
\label{sec:simulation}

The study is designed following the ADEMP structure \citep{morris2019}; results are reported in Section~\ref{sec:results}.

\paragraph{Aims.} (i) To quantify the bias, variance, and coverage of cross-fitted estimators of counterfactual and interventional effects that adjust for a learned representation, as the representation loses confounding information; (ii) to compare the three training regimes; (iii) to test whether reconstruction loss tracks bias (Corollary~\ref{cor:recon}) and whether coverage deteriorates with sample size under insufficiency (Corollary~\ref{cor:coverage}); and (iv) to evaluate diagnostics D1--D4, phenotype-specific estimation, select-then-confirm analysis, and portable representations.

\paragraph{Data-generating mechanisms.} A latent vector $U\sim N_4(0,I)$ generates three modalities: 20 continuous laboratory-type variables $W^{(1)}=\Lambda_1U+e_1$ with $e_1\sim N(0,0.1^2I)$, so that $W$ determines $U$ up to negligible noise and Assumption~\ref{ass:id} holds given $W$; 30 binary diagnosis-type indicators with $\Prob(W^{(2)}_j=1\mid U)=\mathrm{expit}(c_j+\Lambda_{2j}^\top U)$; and 10 four-level ordinal survey-type items obtained by thresholding $\Lambda_3U+e_3$. A rare binary code $R\sim\mathrm{Bernoulli}(0.10)$, independent of $U$, is appended to the diagnosis modality. The causal variables are
\begin{align*}
\Prob(A=1\mid W)&=\mathrm{expit}(\gamma^\top U+\kappa\gamma_RR),\\
\Prob(Z=1\mid A,W)&=\mathrm{expit}(\zeta_0+\zeta_AA+\zeta^\top U+\kappa\zeta_RR),\\
M&=\alpha A+\delta_ZZ+\delta^\top U+\kappa\delta_RR+\epsilon_M,\\
Y&=\vartheta A+\beta M+\lambda_ZZ+\lambda^\top U+\kappa\lambda_RR+\omega AU_1+\epsilon_Y,
\end{align*}
with standard normal errors, and two binary failure modes that are logistic in $(A,U,R)$. The code $R$ raises the probability of treatment, the intermediate confounder, and the mediator, and lowers the outcome ($\gamma_R=\zeta_R=\delta_R=1$, $\lambda_R=-1$), so that the biases of the exposure--mediator and mediator--outcome paths reinforce rather than cancel (Proposition~\ref{prop:linear}). The dial $\kappa\in\{0,0.5,1,2\}$ sets the confounding strength of $R$: with $\kappa=0$ the latent factors are sufficient; with $\kappa>0$ a representation that drops $R$ is not. All coefficients, loadings, and intercepts are fixed across replicates and listed in Supplementary Table~\ref{tab:s1}. Sample sizes are $n\in\{1{,}000,\,3{,}000\}$, and each setting uses 200 replicates, so that the Monte Carlo standard error of estimated coverage is at most 0.035 and about 0.015 near the nominal 95\% level. A single external covariate-only sample of 20{,}000 draws of $W$ is used to train the (R1) encoders, the principal components, and the locked phenotypes, which are three $k$-means clusters of the (R1) representation with four latent dimensions per modality.

\paragraph{Estimands.} The average effect $\psi(1)-\psi(0)$; the phenotype-specific effects $\tau(s)$ for the three locked phenotypes; the interventional indirect and direct effects; and the counterfactual risk of each failure mode under each treatment. True values are computed by Monte Carlo integration over $10^6$ draws of $(U,R)$, using the closed form of $\E\{Y(a,G_{a^*})\mid U,R\}$, in which the intermediate confounder follows its law under $a$ and the mediator its law under $a^*$, as in equation~\eqref{eq:theta}; phenotype-specific truths use $4\times10^5$ draws of $W$ and the locked phenotype map; true counterfactual risks are known for every simulated patient.

\paragraph{Methods.} All methods use five-fold cross-fitting with the same folds and the same nuisance learners, namely logistic and linear regression for adjustment sets with at most 30 columns and cross-validated lasso otherwise, with outcome-type nuisances fitted separately in each exposure arm. They differ only in the adjustment set or in how it is learned: (M1) the oracle, which adjusts for $(U,R)$; (M2) the raw covariates, all 61 variables of $W$; (M3) a VAE with external training (R1), with latent dimension per modality in $\{1,2,4,8\}$, chosen either by validation reconstruction loss or by the nuisance-loss rule D4; (M4) a VAE with fold-specific training (R2); (M5) a VAE with full-sample unsupervised training (R3); (M6) the VAE from (R1) plus key covariates, $\hat\phi^+$, with two laboratory variables and $R$ as the key covariates; (M7) principal components trained on the external sample; and (M8) the VAE from (R3) without cross-fitting of the nuisance models.
Methods (M4)--(M7) use four latent dimensions or components per modality.

\paragraph{Performance measures.} Bias, empirical standard error, root mean squared error, coverage and width of nominal 95\% intervals, and the ratio of mean estimated to empirical standard error, each with Monte Carlo standard errors. For the diagnostics: the share of replicates in which D1 rejects, at $\kappa=0$ and at $\kappa>0$; the area under the curve in D2; the range of estimates across latent dimensions and principal components in D3; and the correlation between reconstruction loss and bias across latent dimensions. For the counterfactual risk profiles: root mean squared error against the true risks and agreement on the failure most likely under each option.

\paragraph{Additional experiments.} Both use $\kappa\in\{0,1\}$ and the two sample sizes. (a) \emph{Select-then-confirm:} 20 candidate mediators, of which two carry non-zero indirect effects; the others are nulls in which the exposure affects the mediator but not the outcome, or the reverse. Same-sample ranking (top three of 20, unadjusted tests) is compared with select-then-confirm (screening on half of the sample, stability by multiplier bootstrap of the influence functions with 200 resamples and a 50\% threshold, Holm-adjusted confirmation of the top three on the other half), on family-wise error, power, and bias of the selected estimates. (b) \emph{Portability:} the portable representation uses the (R1) laboratory and diagnosis encoders only and is applied unchanged to a target population with $U\sim N(0.75\cdot\mathbf 1,I)$ in which the survey modality is unavailable; $\Delta_{\mathrm{port}}$ in the source is compared with the bias in the target. A clustered-data experiment is deferred to a later version.

\section{Simulation results}
\label{sec:results}

\subsection{Representation insufficiency and coverage}

\begin{figure}[htbp]
\centering
\includegraphics[width=\textwidth]{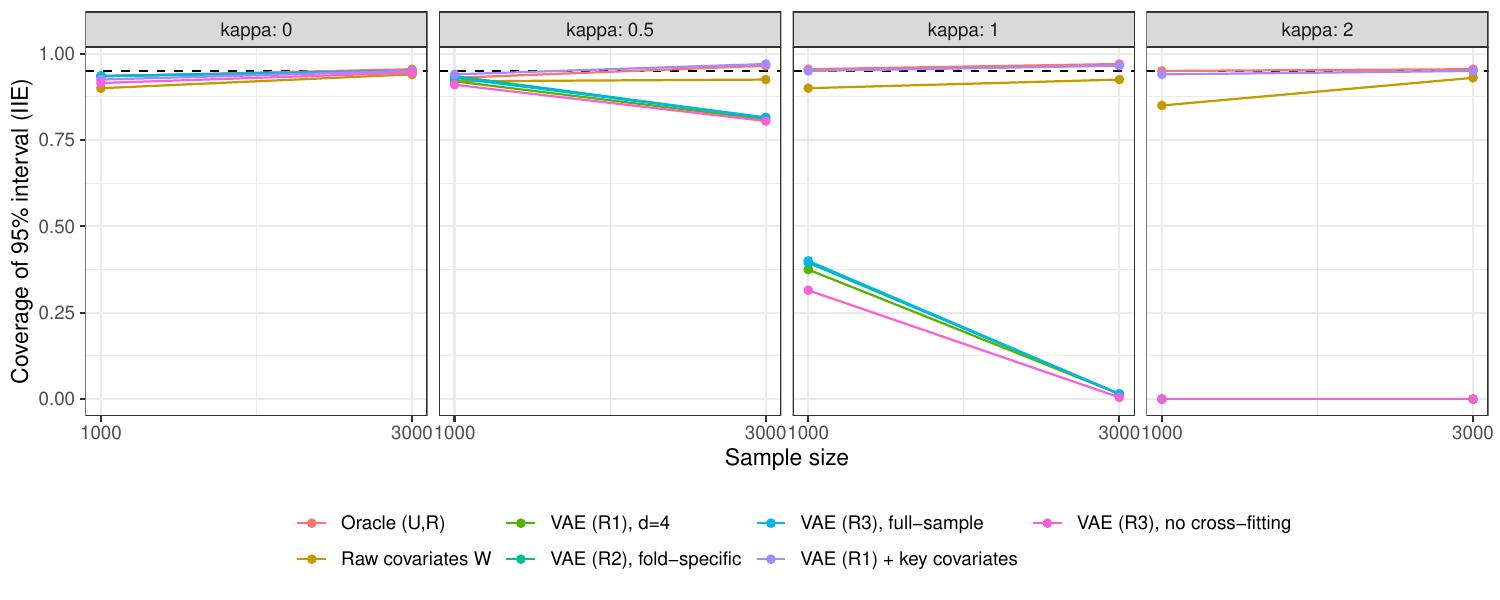}
\caption{Coverage of nominal 95\% intervals for the IIE by sample size, for each confounding strength $\kappa$ of the lost code. The dashed line marks 0.95; 200 replicates per setting.}
\label{fig:coverage}
\end{figure}

With $\kappa=0$, so that the latent factors were sufficient, the VAE representation (R1, $d=4$) estimated the interventional indirect effect with bias $-$0.001 and coverage 0.950 at $n=3,000$, compared with 0.000 and 0.955 for the oracle (Table~\ref{tab:core_IIE}). As the lost code became a stronger confounder, the bias grew to 0.047, 0.186, and 0.491 ($\kappa=0.5$, 1, and 2) and coverage fell to 0.810, 0.015, and 0.000 at $n=3,000$. Where the representation was insufficient, coverage fell as the sample size grew while the bias stayed roughly constant: from 0.920 to 0.810 at $\kappa=0.5$ and from 0.375 to 0.015 at $\kappa=1$ between $n=1,000$ and $n=3,000$, and it was 0.000 at both sample sizes at $\kappa=2$ (Figure~\ref{fig:coverage}), as Corollary~\ref{cor:coverage} predicts. Adding the prespecified key covariates restored coverage (0.965 at $\kappa=1$ for $n=3,000$), as did adjusting for the raw covariates (0.925). The average treatment effect was much less affected: at $\kappa=1$ and $n=3,000$, its bias with the same representation was 0.004, compared with 0.186 for the IIE (Supplementary Table~\ref{tab:core_ATE}), illustrating that the requirements for mediation are stricter than for treatment comparison. The interventional direct effect was biased by a similar amount in the opposite direction ($-$0.181; Supplementary Table~\ref{tab:core_IDE}): the lost code distorted how the effect was split between pathways far more than the effect itself.

\begin{table}[htbp]
\centering
\caption{Estimation of the IIE across methods, confounding strength of the lost code ($\kappa$), and sample size.}
\label{tab:core_IIE}
\small
\begin{tabular}{lrrrrrr}
\toprule
 & \multicolumn{3}{c}{$n=1,000$} & \multicolumn{3}{c}{$n=3,000$} \\
Method & Bias & RMSE & Cov. & Bias & RMSE & Cov. \\
\midrule
\multicolumn{7}{l}{\emph{$\kappa=0$}}\\
\quad Oracle $(U,R)$ & 0.011 & 0.082 & 0.935 & $-$0.000 & 0.046 & 0.955 \\
\quad Raw covariates $W$ & 0.020 & 0.076 & 0.900 & 0.002 & 0.043 & 0.940 \\
\quad VAE (R1), $d=4$ & 0.011 & 0.084 & 0.935 & $-$0.001 & 0.047 & 0.950 \\
\quad VAE (R1), $d$ by nuisance loss & 0.010 & 0.085 & 0.935 & $-$0.001 & 0.047 & 0.950 \\
\quad VAE (R2), fold-specific & 0.008 & 0.085 & 0.935 & $-$0.001 & 0.047 & 0.950 \\
\quad VAE (R3), full-sample & 0.009 & 0.085 & 0.935 & $-$0.002 & 0.047 & 0.950 \\
\quad VAE (R1) $+$ key covariates & 0.011 & 0.086 & 0.925 & $-$0.001 & 0.047 & 0.950 \\
\quad PCA (R1) & 0.010 & 0.084 & 0.945 & $-$0.001 & 0.047 & 0.950 \\
\quad VAE (R3), no cross-fitting & 0.008 & 0.081 & 0.915 & $-$0.002 & 0.047 & 0.945 \\
\multicolumn{7}{l}{\emph{$\kappa=0.5$}}\\
\quad Oracle $(U,R)$ & $-$0.003 & 0.083 & 0.930 & 0.002 & 0.045 & 0.965 \\
\quad Raw covariates $W$ & 0.012 & 0.074 & 0.920 & 0.006 & 0.042 & 0.925 \\
\quad VAE (R1), $d=4$ & 0.043 & 0.095 & 0.920 & 0.047 & 0.065 & 0.810 \\
\quad VAE (R1), $d$ by nuisance loss & 0.040 & 0.093 & 0.930 & 0.047 & 0.064 & 0.810 \\
\quad VAE (R2), fold-specific & 0.041 & 0.093 & 0.935 & 0.047 & 0.064 & 0.815 \\
\quad VAE (R3), full-sample & 0.041 & 0.093 & 0.930 & 0.047 & 0.064 & 0.815 \\
\quad VAE (R1) $+$ key covariates & $-$0.003 & 0.088 & 0.940 & 0.001 & 0.045 & 0.970 \\
\quad PCA (R1) & 0.042 & 0.094 & 0.920 & 0.047 & 0.065 & 0.810 \\
\quad VAE (R3), no cross-fitting & 0.041 & 0.089 & 0.910 & 0.046 & 0.064 & 0.805 \\
\multicolumn{7}{l}{\emph{$\kappa=1$}}\\
\quad Oracle $(U,R)$ & $-$0.004 & 0.081 & 0.955 & 0.005 & 0.041 & 0.970 \\
\quad Raw covariates $W$ & 0.012 & 0.074 & 0.900 & 0.008 & 0.041 & 0.925 \\
\quad VAE (R1), $d=4$ & 0.181 & 0.197 & 0.375 & 0.186 & 0.189 & 0.015 \\
\quad VAE (R1), $d$ by nuisance loss & 0.179 & 0.195 & 0.390 & 0.163 & 0.170 & 0.110 \\
\quad VAE (R2), fold-specific & 0.177 & 0.193 & 0.395 & 0.185 & 0.188 & 0.015 \\
\quad VAE (R3), full-sample & 0.179 & 0.195 & 0.400 & 0.184 & 0.188 & 0.015 \\
\quad VAE (R1) $+$ key covariates & $-$0.003 & 0.085 & 0.950 & 0.005 & 0.043 & 0.965 \\
\quad PCA (R1) & 0.180 & 0.197 & 0.395 & 0.186 & 0.189 & 0.015 \\
\quad VAE (R3), no cross-fitting & 0.175 & 0.190 & 0.315 & 0.183 & 0.187 & 0.005 \\
\multicolumn{7}{l}{\emph{$\kappa=2$}}\\
\quad Oracle $(U,R)$ & 0.010 & 0.090 & 0.950 & $-$0.000 & 0.046 & 0.955 \\
\quad Raw covariates $W$ & 0.029 & 0.082 & 0.850 & 0.004 & 0.043 & 0.930 \\
\quad VAE (R1), $d=4$ & 0.500 & 0.502 & 0.000 & 0.491 & 0.491 & 0.000 \\
\quad VAE (R1), $d$ by nuisance loss & 0.473 & 0.479 & 0.000 & 0.372 & 0.374 & 0.000 \\
\quad VAE (R2), fold-specific & 0.495 & 0.498 & 0.000 & 0.490 & 0.491 & 0.000 \\
\quad VAE (R3), full-sample & 0.496 & 0.499 & 0.000 & 0.490 & 0.490 & 0.000 \\
\quad VAE (R1) $+$ key covariates & 0.009 & 0.098 & 0.940 & $-$0.000 & 0.046 & 0.950 \\
\quad PCA (R1) & 0.501 & 0.503 & 0.000 & 0.491 & 0.492 & 0.000 \\
\quad VAE (R3), no cross-fitting & 0.493 & 0.495 & 0.000 & 0.488 & 0.488 & 0.000 \\
\bottomrule
\end{tabular}

\vspace{0.15cm}
\begin{minipage}{0.95\textwidth}\footnotesize
\textit{Note.} 200 replicates per setting; Cov., coverage of nominal 95\% intervals. Largest Monte Carlo standard errors: bias 0.007, coverage 0.035.
\end{minipage}
\end{table}

\subsection{Reconstruction loss does not track bias}

\begin{figure}[htbp]
\centering
\includegraphics[width=0.7\textwidth]{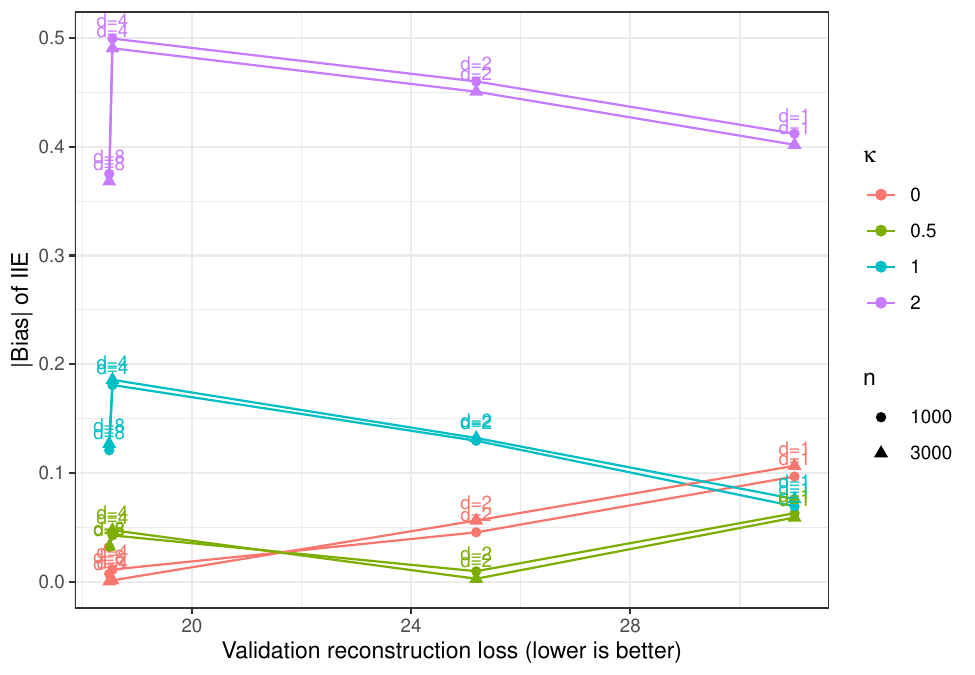}
\caption{Absolute bias of the IIE against validation reconstruction loss across latent dimensions $d$ of the (R1) VAE, by confounding strength $\kappa$ and sample size $n$.}
\label{fig:recon}
\end{figure}

Reconstruction loss was lowest at $d=8$ (Table~\ref{tab:dsweep}), so the reconstruction rule selected that dimension in every replicate. Across latent dimensions, the correlation between reconstruction loss and the absolute bias of the IIE ranged from $-$0.82 to 1.00 across settings (Figure~\ref{fig:recon}): it was positive at $\kappa=0$ (0.99 and 1.00 at $n=1,000$ and $3,000$) but negative at $\kappa=1$ ($-$0.81 and $-$0.82), where better reconstruction went with larger bias. At $\kappa=1$ and $n=3,000$, the dimension chosen by reconstruction gave bias 0.127 and coverage 0.185, and the dimension chosen by nuisance loss gave 0.163 and 0.110. At $\kappa\ge1$ every candidate dimension gave a biased estimate of the IIE, so neither rule could remove the bias; what detected the problem was D1 (Section~\ref{sec:res_diag}), and what removed it was adding the key covariates.

\begin{table}[htbp]
\centering
\caption{Latent dimension, reconstruction loss, and estimation of the IIE (regime R1). ``Chosen'' is the share of replicates in which the nuisance-loss rule D4 selected that dimension; the reconstruction rule selects the dimension with the lowest reconstruction loss in every replicate.}
\label{tab:dsweep}
\small
\begin{tabular}{rrrrrrrrr}
\toprule
 & & & \multicolumn{3}{c}{$n=1,000$} & \multicolumn{3}{c}{$n=3,000$} \\
$\kappa$ & $d$ & Recon. loss & Bias & Cov. & Chosen & Bias & Cov. & Chosen \\
\midrule
0 & 1 & 31.00 & $-$0.097 & 0.830 & 0\% & $-$0.107 & 0.515 & 0\% \\
0 & 2 & 25.19 & $-$0.045 & 0.920 & 2\% & $-$0.056 & 0.805 & 0\% \\
0 & 4 & 18.55 & 0.011 & 0.935 & 98\% & $-$0.001 & 0.950 & 100\% \\
0 & 8 & 18.49 & 0.007 & 0.940 & 0\% & $-$0.000 & 0.945 & 0\% \\
0.5 & 1 & 31.00 & $-$0.063 & 0.880 & 0\% & $-$0.059 & 0.800 & 0\% \\
0.5 & 2 & 25.19 & $-$0.010 & 0.935 & 4\% & $-$0.003 & 0.940 & 0\% \\
0.5 & 4 & 18.55 & 0.043 & 0.920 & 96\% & 0.047 & 0.810 & 97\% \\
0.5 & 8 & 18.49 & 0.032 & 0.950 & 0\% & 0.032 & 0.885 & 3\% \\
1 & 1 & 31.00 & 0.069 & 0.885 & 0\% & 0.076 & 0.700 & 0\% \\
1 & 2 & 25.19 & 0.130 & 0.630 & 3\% & 0.132 & 0.195 & 0\% \\
1 & 4 & 18.55 & 0.181 & 0.375 & 97\% & 0.186 & 0.015 & 65\% \\
1 & 8 & 18.49 & 0.121 & 0.705 & 0\% & 0.127 & 0.185 & 35\% \\
2 & 1 & 31.00 & 0.412 & 0.010 & 0\% & 0.402 & 0.000 & 0\% \\
2 & 2 & 25.19 & 0.460 & 0.005 & 4\% & 0.451 & 0.000 & 0\% \\
2 & 4 & 18.55 & 0.500 & 0.000 & 78\% & 0.491 & 0.000 & 3\% \\
2 & 8 & 18.49 & 0.375 & 0.005 & 19\% & 0.368 & 0.000 & 97\% \\
\bottomrule
\end{tabular}
\end{table}

\subsection{Training regimes}

At $\kappa=0$, the three training regimes gave similar results: IIE coverage at $n=3,000$ was 0.950 for external training, 0.950 for fold-specific training, and 0.950 for full-sample unsupervised training (Table~\ref{tab:core_IIE}). Omitting cross-fitting of the nuisance models gave coverage 0.945 at $n=3,000$ and 0.915 at $n=1,000$. At $\kappa>0$ the three regimes and the principal components were biased to the same degree (IIE bias 0.184 to 0.186 at $\kappa=1$ and $n=3,000$), so the bias came from the lost code rather than from how the encoder was trained.

\subsection{Diagnostics}
\label{sec:res_diag}

Diagnostic D1 flagged the $d=4$ representation in 0\% of replicates at $\kappa=0$ and $n=3,000$, where it was sufficient, and in 100\% at $\kappa=1$; it flagged the $d=1$ representation, which cannot recover the four latent factors, in 100\% of replicates at $\kappa=0$ (Table~\ref{tab:diagnostics}). At $\kappa=0.5$, where the bias was smaller, D1 flagged the $d=4$ representation in 28\% of replicates at $n=1,000$ and in 99\% at $n=3,000$. The cross-validated area under the curve for predicting the exposure from the representation ranged from 0.668 to 0.683, so exposure leakage was not a concern in this design. Estimates varied across latent dimensions by 0.112 on average at $\kappa=1$ and $n=3,000$ (D3), compared with 0.110 at $\kappa=0$, where the spread came from the lower-dimensional representations that cannot recover $U$.

\begin{table}[htbp]
\centering
\caption{Diagnostics. D1: share of replicates in which adding the raw covariates significantly improved a nuisance model (Holm-adjusted, 5\% level). D2: cross-validated area under the curve for predicting $A$. D3: mean range of IIE estimates across latent dimensions and PCA within a replicate.}
\label{tab:diagnostics}
\small
\begin{tabular}{rrrrrrr}
\toprule
$\kappa$ & $n$ & D1, $d=1$ & D1, $d=4$ & D2 AUC, VAE & D2 AUC, raw & D3 range \\
\midrule
0 & 1,000 & 100\% & 2\% & 0.678 & 0.683 & 0.127 \\
0 & 3,000 & 100\% & 0\% & 0.683 & 0.685 & 0.110 \\
0.5 & 1,000 & 100\% & 28\% & 0.671 & 0.678 & 0.123 \\
0.5 & 3,000 & 100\% & 99\% & 0.683 & 0.687 & 0.109 \\
1 & 1,000 & 100\% & 100\% & 0.670 & 0.685 & 0.127 \\
1 & 3,000 & 100\% & 100\% & 0.681 & 0.696 & 0.112 \\
2 & 1,000 & 100\% & 100\% & 0.668 & 0.711 & 0.137 \\
2 & 3,000 & 100\% & 100\% & 0.675 & 0.716 & 0.125 \\
\bottomrule
\end{tabular}
\end{table}

\subsection{Counterfactual comparisons, phenotypes, pathway selection, and portability}

Phenotype-specific effects were covered at an average rate of 0.955 with the $d=4$ representation at $\kappa=0$ and 0.947 at $\kappa=1$ ($n=3,000$; Table~\ref{tab:counterfactual}). Patient-level counterfactual risk profiles were more sensitive: their root mean squared error against the true risks rose from 0.046 at $\kappa=0$ to 0.089 and 0.137 at $\kappa=1$ and 2, compared with 0.051 when the key covariates were added, and agreement with the true most likely failure fell from 90\% to 85\% and 82\% of patients and options.

In the pathway-selection experiment (Supplementary Table~\ref{tab:stc}), same-sample ranking declared at least one null pathway significant in 52\% and 56\% of replicates at $\kappa=0$, whereas select-then-confirm declared none, with power 47\% at $n=1,000$ and 99\% at $n=3,000$. At $\kappa=1$ the family-wise error of select-then-confirm rose to 21\% and 34\%, and the selection bias was 0.32 to 0.36 under both procedures: sample splitting removes the bias from ranking, not the bias from the representation.

Restricting the representation to the shared modalities changed the source estimate of the IIE by 0.000 on average, with an interval containing zero in 95\% to 97\% of replicates (Supplementary Table~\ref{tab:port}). In the shifted target population, the portable representation was nearly unbiased at $\kappa=0$ (bias 0.001 to 0.003) but had bias of about 0.18 at $\kappa=1$, the same as in the source: a value of $\Delta_{\mathrm{port}}$ near zero shows that the restriction loses nothing, not that the representation is sufficient.

\begin{table}[htbp]
\centering
\caption{Counterfactual comparisons: average effect (bias, coverage), phenotype-specific effects (mean coverage across phenotypes), counterfactual risk profiles (root mean squared error against the true risks), and agreement on which failure is most likely under each option.}
\label{tab:counterfactual}
\small
\resizebox{\textwidth}{!}{%
\begin{tabular}{rlrrrrrrrrrr}
\toprule
 & & \multicolumn{5}{c}{$n=1,000$} & \multicolumn{5}{c}{$n=3,000$} \\
$\kappa$ & Method & ATE bias & ATE cov. & $\tau(s)$ cov. & Risk RMSE & Agree & ATE bias & ATE cov. & $\tau(s)$ cov. & Risk RMSE & Agree \\
\midrule
0 & Oracle $(U,R)$ & 0.014 & 0.960 & 0.955 & 0.053 & 0.893 & $-$0.001 & 0.935 & 0.953 & 0.030 & 0.936 \\
0 & Raw covariates $W$ & 0.033 & 0.950 & 0.942 & 0.066 & 0.834 & 0.006 & 0.930 & 0.953 & 0.042 & 0.906 \\
0 & VAE (R1), $d=4$ & 0.016 & 0.975 & 0.953 & 0.079 & 0.841 & 0.000 & 0.930 & 0.955 & 0.046 & 0.901 \\
0 & VAE (R1) $+$ key covariates & 0.017 & 0.970 & 0.957 & 0.089 & 0.826 & 0.000 & 0.930 & 0.955 & 0.051 & 0.891 \\
0.5 & Oracle $(U,R)$ & $-$0.012 & 0.955 & 0.948 & 0.053 & 0.901 & 0.001 & 0.940 & 0.960 & 0.030 & 0.943 \\
0.5 & Raw covariates $W$ & 0.006 & 0.930 & 0.948 & 0.075 & 0.818 & 0.007 & 0.935 & 0.952 & 0.047 & 0.899 \\
0.5 & VAE (R1), $d=4$ & $-$0.010 & 0.960 & 0.950 & 0.089 & 0.818 & 0.002 & 0.965 & 0.953 & 0.060 & 0.874 \\
0.5 & VAE (R1) $+$ key covariates & $-$0.010 & 0.950 & 0.952 & 0.089 & 0.837 & 0.002 & 0.945 & 0.958 & 0.050 & 0.902 \\
1 & Oracle $(U,R)$ & 0.001 & 0.940 & 0.948 & 0.053 & 0.908 & 0.003 & 0.975 & 0.953 & 0.031 & 0.945 \\
1 & Raw covariates $W$ & 0.018 & 0.950 & 0.938 & 0.079 & 0.838 & 0.009 & 0.945 & 0.948 & 0.047 & 0.910 \\
1 & VAE (R1), $d=4$ & 0.002 & 0.960 & 0.953 & 0.111 & 0.807 & 0.004 & 0.970 & 0.947 & 0.089 & 0.850 \\
1 & VAE (R1) $+$ key covariates & 0.006 & 0.955 & 0.948 & 0.089 & 0.846 & 0.005 & 0.975 & 0.950 & 0.051 & 0.904 \\
2 & Oracle $(U,R)$ & 0.015 & 0.970 & 0.955 & 0.053 & 0.906 & 0.001 & 0.930 & 0.947 & 0.031 & 0.943 \\
2 & Raw covariates $W$ & 0.030 & 0.935 & 0.935 & 0.084 & 0.843 & 0.005 & 0.925 & 0.937 & 0.049 & 0.910 \\
2 & VAE (R1), $d=4$ & $-$0.000 & 0.955 & 0.952 & 0.155 & 0.787 & $-$0.016 & 0.930 & 0.950 & 0.137 & 0.821 \\
2 & VAE (R1) $+$ key covariates & 0.019 & 0.955 & 0.953 & 0.090 & 0.841 & 0.002 & 0.925 & 0.943 & 0.051 & 0.903 \\
\bottomrule
\end{tabular}}
\end{table}

\section{Discussion}

Learned representations make high-dimensional, multimodal adjustment practical, but they move part of the identification problem into the representation. The central message of this paper is that the error this creates is first order. Cross-fitting and double robustness protect against slow estimation of nuisance functions given the representation; they do not protect against information the representation has discarded, and that bias does not shrink with sample size. Larger samples make it more visible, because intervals narrow around the wrong value.

Three consequences for practice follow. First, the requirements depend on the estimand: comparing treatments requires a representation that is balancing or prognostic, whereas interventional mediation requires that the representation preserve the information used by the exposure, intermediate-confounder, mediator, and outcome models together. A representation validated for one purpose should not be assumed adequate for the other. Second, neither reconstruction quality nor any other covariate-only criterion, including physics-informed penalties on the covariates, is evidence of adequacy for causal analysis; representations should be chosen by the losses of the nuisance functions the estimand requires and checked with diagnostics such as D1--D3. Third, representations should be trained outside the estimation sample, either on external covariate data or within cross-fitting folds, and known strong confounders should be added to the representation rather than left to it. Phenotypes and selected pathways should be locked on data separate from the data used to estimate their effects.

The work has limitations. All results assume no unmeasured confounding given $W$; the diagnostics detect predictive insufficiency of the representation relative to measured covariates, not confounding by unmeasured variables, and a non-significant diagnostic is not proof of sufficiency. The closed-form bias is derived for a linear Gaussian model; the decomposition in Theorem~\ref{thm:decomp} is general but conditional on a representation trained independently of the estimation sample, and full-sample unsupervised training requires an additional stability condition. The mediation estimator uses a single binary intermediate confounder. Counterfactual risk profiles are plug-in quantities and inherit the calibration of the outcome models. Mechanism-informed objectives that link the exposure, mediator, and outcome, and digital twins built on them, may reduce representation bias; whether they do is an empirical question that the diagnostics proposed here can answer, and we leave it to future work. Finally, the simulation study uses one data-generating design, two sample sizes, and 200 replicates per setting; applications to obesity treatment, comparing response and failure across metabolic and bariatric surgery and glucagon-like peptide-1 receptor agonist therapy and decomposing the role of early weight change, are reported separately.

\section*{Data and Code Availability}

No patient data were analyzed; all results are based on simulated data. Code implementing the estimators, the diagnostics, and the simulation study is publicly available at \url{https://github.com/congca0/reconstruction-is-not-enough} and archived at \url{https://doi.org/10.5281/zenodo.23524452}.

\section*{Ethics Statement}
This study used only simulated data. No human participants or identifiable data were involved,
and ethical approval was not required.
\section*{Funding}

The author received no specific funding for this work.

\section*{Competing Interests}

The author declares no competing interests.

\appendix

\section{Proofs}
\label{app:proofs}

\paragraph{Proposition~\ref{prop:id}(a).} If $\phi$ is balancing, $g(a\mid W)=g(a\mid\phi(W))$. By Assumption~\ref{ass:id}(ii), $\Prob\{A=a\mid Y(a),\phi\}=\E[\Prob\{A=a\mid Y(a),W\}\mid Y(a),\phi]=\E\{g(a\mid W)\mid Y(a),\phi\}=g(a\mid\phi)$, so $Y(a)\indep A\mid\phi(W)$ and $\mu_a(\phi)=\E\{Y(a)\mid\phi\}$; taking expectations gives $\psi_\phi(a)=\psi_W(a)$. If $\phi$ is prognostic, $\mu_a(W)=f_a\{\phi(W)\}$ and $\E(Y\mid A=a,\phi)=\E\{\mu_a(W)\mid A=a,\phi\}=f_a(\phi)$, so $\psi_\phi(a)=\E\{f_a(\phi)\}=\E\{\mu_a(W)\}=\psi_W(a)$. When $S$ is a function of $\phi(W)$, the same arguments hold conditionally on $S$ by the tower property.

\paragraph{Proposition~\ref{prop:id}(b).} The $\phi$-adjusted nuisances are conditional averages of the $W$-adjusted ones: $\E(Y\mid a,z,m,\phi)=\E\{b(a,z,m,W)\mid a,z,m,\phi\}$, $\Prob(Z=z\mid a,\phi)=\E\{q(z\mid a,W)\mid A=a,\phi\}$, and the density of $M$ given $(A,\phi)=(a^*,\phi)$ is $\E\{\bar p(m\mid a^*,W)\mid A=a^*,\phi\}$. If each $W$-adjusted nuisance depends on $W$ only through $\phi(W)$, each conditional average equals the function itself evaluated at $\phi(W)$. Substituting into the $\phi$-version of \eqref{eq:theta} and taking the outer expectation over $\phi(W)$ gives \eqref{eq:theta}.

\paragraph{Proposition~\ref{prop:id}(c).} Process sufficiency implies $A\indep W\mid\phi$ and, by weak union, $M\indep W\mid(A,Z,\phi)$ and $Z\indep W\mid(A,\phi)$. The argument of part (a) applied to $Y(a,m)$ and to $M(a^*)$ gives Assumption~\ref{ass:id}(ii) given $\phi$. For (iii), $\Prob\{M\mid Y(a,m),A,Z,\phi\}=\E[\Prob\{M\mid A,Z,W\}\mid Y(a,m),A,Z,\phi]=\Prob(M\mid A,Z,\phi)$. Positivity transfers because $g(a\mid\phi)=\E\{g(a\mid W)\mid\phi\}$. The law of $M(a^*)$ given $W$ is that of $M$ given $(A=a^*,W)$, which by $M\indep W\mid(A,\phi)$ equals that of $M$ given $(A=a^*,\phi)$, the law of $M(a^*)$ given $\phi$. The nuisances in (b) are then $\phi$-measurable, so (b) applies.

\paragraph{Proposition~\ref{prop:id}(d).} Let $A$ be randomized, $\phi$ constant, and $W$ a binary confounder of $M$ and $Y$ that is unaffected by $A$. Then $\phi$ is balancing and $\psi_\phi(a)=\E(Y\mid A=a)=\psi_W(a)$, while $\theta_\phi(a,a^*)$ uses $\E(Y\mid a,z,m)$ in place of $\E\{b(a,z,m,W)\}$ and differs from $\theta_W(a,a^*)$ whenever $W$ affects both $M$ and $Y$.

\paragraph{Proposition~\ref{prop:covonly}.} Because $W$ takes values in a standard Borel space and is not almost surely a function of $\phi(W)$, there is a bounded measurable function $r$ with $\E[\mathrm{Var}\{r(W)\mid\phi(W)\}]>0$; otherwise every bounded function of $W$ would be almost surely equal to a function of $\phi(W)$, and $W$ would be almost surely a function of $\phi(W)$ by the Doob--Dynkin lemma. Let $\rho(W)=r(W)-\E\{r(W)\mid\phi(W)\}$, so that $\E(\rho\mid\phi)=0$, $|\rho|\le c$ for some $c<\infty$, and $\E(\rho^2)>0$. Fix the law of $W$, let $\epsilon_M,\epsilon_Y\sim N(0,1)$ be independent of each other and of $W$ and $A$, and let $Z$ be degenerate. Under $P_0$, $A\sim\mathrm{Bernoulli}(1/2)$ independently of $W$, $M=A+\epsilon_M$, and $Y=M+\epsilon_Y$; then $(A,Z,M)\indep W$, so $\phi$ is process-sufficient. Under $P_1$, $\Prob(A=1\mid W)=1/2+t\rho(W)$ with $0<t<1/(2c)$, $M=A+\rho(W)+\epsilon_M$, and $Y=A+M+\rho(W)+\epsilon_Y$. Then $\psi_W(1)=2$, whereas $\E(\rho\mid A=1,\phi)=\E\{\rho(1/2+t\rho)\mid\phi\}/\E\{1/2+t\rho\mid\phi\}=2t\,\E(\rho^2\mid\phi)$, so $\psi_\phi(1)=2+4t\,\E(\rho^2)\neq2$. Under $P_2$, $A\sim\mathrm{Bernoulli}(1/2)$ independently of $W$, $M=A+\rho(W)+\epsilon_M$, and $Y=M+\rho(W)+\epsilon_Y$. Then $b(a,m,W)=m+\rho(W)$ and the draw of $M(a^*)$ given $W$ has mean $a^*+\rho(W)$, so $\theta_W(a,a^*)=a^*$ and $\IIE_W=1$. Given $\phi$, let $S=\rho+\epsilon_M$ and $h_\phi(s)=\E(\rho\mid S=s,\phi)$; because $\epsilon_M$ is Gaussian, $h_\phi$ is nondecreasing, with derivative $\mathrm{Var}(\rho\mid S=s,\phi)$, and strictly increasing on the event $\{\mathrm{Var}(\rho\mid\phi)>0\}$, which has positive probability. Since $A$ is independent of $(W,\epsilon_M)$, $\E(Y\mid A=a,M=m,\phi)=m+h_\phi(m-a)$, and $M$ given $(A=a^*,\phi)$ is distributed as $a^*+S$. Hence $\theta_\phi(1,1)=1+\E\{h_\phi(S)\}=1+\E(\rho)=1$ and $\theta_\phi(1,0)=\E\{h_\phi(S-1)\}<\E\{h_\phi(S)\}=0$, so $\IIE_\phi>1=\IIE_W$. In all three laws every dependence among $(A,M,Y)$ operates through $W$ and the error terms are independent, so Assumption~\ref{ass:id} holds, with positivity guaranteed by $0<t<1/(2c)$ and by the Gaussian errors. A covariate-only objective depends on the law only through the common marginal law of $W$, so it takes the same value at $\phi$ under $P_0$, $P_1$, and $P_2$.

\paragraph{Proposition~\ref{prop:linear}.} All variables are jointly Gaussian. Since $R\indep U$, $\mathrm{Cov}(A,R\mid U)=\gamma_R\sigma_R^2$ and $\mathrm{Var}(A\mid U)=\gamma_R^2\sigma_R^2+\sigma_A^2$, so $\E(R\mid A,U)=\kappa(A-\gamma^\top U)$ and the coefficient of $A$ in $\E(M\mid A,U)$ is $\alpha+\delta_R\kappa$. Next, $\mathrm{Var}(R\mid A,U)=v_R$, $\mathrm{Cov}(M,R\mid A,U)=\delta_Rv_R$, and $\mathrm{Var}(M\mid A,U)=\delta_R^2v_R+\sigma_M^2$, so the coefficient of $M$ in $\E(Y\mid A,M,U)=\vartheta A+\beta M+\lambda^\top U+\lambda_R\E(R\mid A,M,U)$ is $\tilde\beta$. With linear Gaussian nuisances, the $\phi$-adjusted functional gives $\IIE_\phi(a,a^*)=\tilde\beta\{\E(M\mid a,U)-\E(M\mid a^*,U)\}=\tilde\alpha\tilde\beta(a-a^*)$. The total-effect coefficient in $\E(Y\mid A,U)$ is $\vartheta+\beta\alpha+(\lambda_R+\beta\delta_R)\kappa$.

\paragraph{Corollary~\ref{cor:recon}.} The best squared-error reconstruction of $R$ from $U$ is $\E(R\mid U)=0$, with error $\sigma_R^2$; the other coordinates are unaffected. Fix $\beta\neq0$, $\lambda_R=0$, $\sigma_R^2=\varepsilon$, and $\gamma_R=\sigma_A/\sigma_R$, so that $\kappa=\sigma_R/(2\sigma_A)$ and $\tilde\beta=\beta$. The bias is $\beta\delta_R\sigma_R/(2\sigma_A)$, which exceeds $B$ in absolute value for $|\delta_R|>2B\sigma_A/(|\beta|\sigma_R)$.

\paragraph{Theorem~\ref{thm:decomp}.} Conditionally on $\hat\phi$, the adjustment set $\hat\phi(W)$ is a fixed function of $W$, and the standard expansion for cross-fitted one-step estimators applies \citep{chernozhukov2018,kennedy2016,diaz2021}: $\hat\Psi-\Psi_{\hat\phi}=(\mathbb P_n-\mathbb P)D_{\hat\phi}+R_2(\hat\eta,\eta_{\hat\phi})+(\mathbb P_n-\mathbb P)(\hat D-D_{\hat\phi})$, where the last term is $o_{\mathbb P}(n^{-1/2})$ by cross-fitting and $L_2$-consistency. Adding and subtracting $\Psi_W$ gives the result. For $\psi(a)$, the augmented inverse-probability-weighted estimator has $R_2=\E[\{\hat\mu_a-\mu_a\}\{\hat g-g\}/\hat g]$, with all functions evaluated at $\hat\phi(W)$.

\paragraph{Corollary~\ref{cor:coverage}.} By Theorem~\ref{thm:decomp}, $\sqrt n(\hat\Psi-\Psi_W)/\hat\sigma=\sqrt n\,b/\sigma+N(0,1)+o_{\mathbb P}(1)$, and the coverage of $\hat\Psi\pm1.96\,\hat\sigma/\sqrt n$ follows.

\section{Influence function used for estimation}
\label{app:eif}

With $a'$ and $a^*$ fixed, write $g(a\mid V)$, $q(z\mid a,V)$, $e(a\mid M,V)=\Prob(A=a\mid M,V)$, $h(z\mid a,M,V)=\Prob(Z=z\mid A=a,M,V)$, and $b(a,z,m,V)$, and define $u(z,V)=\E\{b(a',z,M,V)\mid A=a^*,V\}$, $v(m,V)=\sum_zb(a',z,m,V)\,q(z\mid a',V)$, and $\Phi(V)=\sum_zu(z,V)\,q(z\mid a',V)$. The efficient influence function of $\theta_V(a',a^*)$ \citep{diaz2021} is
\begin{align*}
D={}&\frac{\mathbb 1(A=a')\,e(a^*\mid M,V)\,q(Z\mid a',V)}{g(a^*\mid V)\,e(a'\mid M,V)\,h(Z\mid a',M,V)}\{Y-b(a',Z,M,V)\}\\
&+\frac{\mathbb 1(A=a')}{g(a'\mid V)}\{u(Z,V)-\Phi(V)\}+\frac{\mathbb 1(A=a^*)}{g(a^*\mid V)}\{v(M,V)-\Phi(V)\}+\Phi(V)-\theta_V(a',a^*),
\end{align*}
where the first weight is $\bar p(M\mid a^*,V)/\{g(a'\mid V)\,p(M\mid a',Z,V)\}$ rewritten by Bayes' rule. The one-step estimator is the sample mean of $D+\theta_V(a',a^*)$ with cross-fitted nuisance estimates; $u$ is estimated by regressing $\hat b(a',z,M,V)$ on $V$ and $\hat q(1\mid a^*,V)$ among observations with $A=a^*$.

\newpage
\section{Supplementary tables}
\setcounter{table}{0}
\renewcommand{\thetable}{S\arabic{table}}

\begin{table}[htbp]
\centering
\caption{Fixed parameters of the data-generating mechanism.}
\label{tab:s1}
\small
\begin{tabular}{lp{11cm}}
\toprule
Component & Specification \\
\midrule
Latent factors & $U\sim N_4(0,I)$ \\
Laboratory modality & 20 variables, $W^{(1)}=\Lambda_1U+e_1$, $e_1\sim N(0,0.1^2I)$ \\
Diagnosis modality & 30 binary codes, logistic in $U$, plus $R\sim\mathrm{Bernoulli}(0.1)$ \\
Survey modality & 10 four-level items, thresholded $\Lambda_3U+e_3$, $e_3\sim N(0,0.7^2I)$ \\
Loadings & entries $N(0,0.8^2)$, drawn once (seed 20261020); diagnosis intercepts $U(-2,-0.5)$ \\
Exposure $A$ & logit: $\gamma=(0.5,\ -0.4,\ 0.3,\ 0.2)$, $\gamma_R=1$ \\
Intermediate confounder $Z$ & logit: $\zeta_0=-0.3$, $\zeta_A=0.8$, $\zeta=(0.3,\ 0.3,\ -0.2,\ 0)$, $\zeta_R=1$ \\
Mediator $M$ & $\alpha=-1$, $\delta_Z=0.5$, $\delta=(0.4,\ 0.3,\ -0.3,\ 0.2)$, $\delta_R=1$, $\epsilon_M\sim N(0,1)$ \\
Outcome $Y$ & $\vartheta=0.5$, $\beta=0.8$, $\lambda_Z=0.4$, $\lambda=(0.5,\ -0.3,\ 0.2,\ 0.3)$, $\lambda_R=-1$, $\omega=0.4$ \\
Failure mode 1 & logit: $-1.5+0.6A+(0.5,\ 0.2,\ -0.3,\ 0)^\top U+\kappa(1.5)R$ \\
Failure mode 2 & logit: $-1-0.5A+(-0.2,\ 0.4,\ 0.3,\ 0.2)^\top U+\kappa(-1.5)R$ \\
Dial and sizes & $\kappa\in\{0, 0.5, 1, 2\}$; $n\in\{1{,}000, 3{,}000\}$; 200 replicates \\
Representations & VAE latent dimension per modality 1, 2, 4, 8 for (R1), 4 otherwise; external sample 20,000; 5-fold cross-fitting \\
Pathway selection & 20 candidate mediators, two with non-zero indirect effects \\
Portability & target $U\sim N(0.75\mathbf 1,I)$, survey modality unavailable \\
\bottomrule
\end{tabular}
\end{table}

\begin{table}[htbp]
\centering
\caption{Estimation of the ATE across methods, confounding strength of the lost code ($\kappa$), and sample size.}
\label{tab:core_ATE}
\small
\begin{tabular}{lrrrrrr}
\toprule
 & \multicolumn{3}{c}{$n=1,000$} & \multicolumn{3}{c}{$n=3,000$} \\
Method & Bias & RMSE & Cov. & Bias & RMSE & Cov. \\
\midrule
\multicolumn{7}{l}{\emph{$\kappa=0$}}\\
\quad Oracle $(U,R)$ & 0.014 & 0.092 & 0.960 & $-$0.001 & 0.053 & 0.935 \\
\quad Raw covariates $W$ & 0.033 & 0.095 & 0.950 & 0.006 & 0.052 & 0.930 \\
\quad VAE (R1), $d=4$ & 0.016 & 0.093 & 0.975 & 0.000 & 0.052 & 0.930 \\
\quad VAE (R1), $d$ by nuisance loss & 0.017 & 0.094 & 0.975 & 0.000 & 0.052 & 0.930 \\
\quad VAE (R2), fold-specific & 0.018 & 0.094 & 0.975 & 0.001 & 0.052 & 0.935 \\
\quad VAE (R3), full-sample & 0.016 & 0.093 & 0.965 & 0.001 & 0.053 & 0.930 \\
\quad VAE (R1) $+$ key covariates & 0.017 & 0.094 & 0.970 & 0.000 & 0.053 & 0.930 \\
\quad PCA (R1) & 0.014 & 0.093 & 0.985 & $-$0.001 & 0.052 & 0.935 \\
\quad VAE (R3), no cross-fitting & 0.016 & 0.092 & 0.975 & 0.002 & 0.052 & 0.925 \\
\multicolumn{7}{l}{\emph{$\kappa=0.5$}}\\
\quad Oracle $(U,R)$ & $-$0.012 & 0.096 & 0.955 & 0.001 & 0.052 & 0.940 \\
\quad Raw covariates $W$ & 0.006 & 0.095 & 0.930 & 0.007 & 0.052 & 0.935 \\
\quad VAE (R1), $d=4$ & $-$0.010 & 0.097 & 0.960 & 0.002 & 0.052 & 0.965 \\
\quad VAE (R1), $d$ by nuisance loss & $-$0.007 & 0.098 & 0.960 & 0.002 & 0.052 & 0.960 \\
\quad VAE (R2), fold-specific & $-$0.008 & 0.097 & 0.950 & 0.003 & 0.052 & 0.955 \\
\quad VAE (R3), full-sample & $-$0.010 & 0.098 & 0.955 & 0.003 & 0.053 & 0.950 \\
\quad VAE (R1) $+$ key covariates & $-$0.010 & 0.098 & 0.950 & 0.002 & 0.052 & 0.945 \\
\quad PCA (R1) & $-$0.011 & 0.098 & 0.955 & 0.000 & 0.052 & 0.950 \\
\quad VAE (R3), no cross-fitting & $-$0.010 & 0.095 & 0.945 & 0.003 & 0.053 & 0.940 \\
\multicolumn{7}{l}{\emph{$\kappa=1$}}\\
\quad Oracle $(U,R)$ & 0.001 & 0.097 & 0.940 & 0.003 & 0.052 & 0.975 \\
\quad Raw covariates $W$ & 0.018 & 0.094 & 0.950 & 0.009 & 0.052 & 0.945 \\
\quad VAE (R1), $d=4$ & 0.002 & 0.094 & 0.960 & 0.004 & 0.053 & 0.970 \\
\quad VAE (R1), $d$ by nuisance loss & 0.006 & 0.098 & 0.950 & 0.005 & 0.053 & 0.965 \\
\quad VAE (R2), fold-specific & 0.003 & 0.094 & 0.965 & 0.005 & 0.052 & 0.965 \\
\quad VAE (R3), full-sample & 0.003 & 0.095 & 0.950 & 0.005 & 0.052 & 0.970 \\
\quad VAE (R1) $+$ key covariates & 0.006 & 0.098 & 0.955 & 0.005 & 0.053 & 0.975 \\
\quad PCA (R1) & 0.001 & 0.094 & 0.965 & 0.003 & 0.052 & 0.970 \\
\quad VAE (R3), no cross-fitting & 0.002 & 0.092 & 0.940 & 0.005 & 0.052 & 0.970 \\
\multicolumn{7}{l}{\emph{$\kappa=2$}}\\
\quad Oracle $(U,R)$ & 0.015 & 0.104 & 0.970 & 0.001 & 0.060 & 0.930 \\
\quad Raw covariates $W$ & 0.030 & 0.099 & 0.935 & 0.005 & 0.057 & 0.925 \\
\quad VAE (R1), $d=4$ & $-$0.000 & 0.093 & 0.955 & $-$0.016 & 0.057 & 0.930 \\
\quad VAE (R1), $d$ by nuisance loss & 0.005 & 0.098 & 0.955 & $-$0.012 & 0.056 & 0.920 \\
\quad VAE (R2), fold-specific & 0.004 & 0.093 & 0.960 & $-$0.015 & 0.056 & 0.935 \\
\quad VAE (R3), full-sample & 0.002 & 0.092 & 0.950 & $-$0.015 & 0.056 & 0.940 \\
\quad VAE (R1) $+$ key covariates & 0.019 & 0.111 & 0.955 & 0.002 & 0.062 & 0.925 \\
\quad PCA (R1) & 0.001 & 0.093 & 0.955 & $-$0.018 & 0.058 & 0.935 \\
\quad VAE (R3), no cross-fitting & 0.003 & 0.092 & 0.955 & $-$0.015 & 0.056 & 0.945 \\
\bottomrule
\end{tabular}

\vspace{0.15cm}
\begin{minipage}{0.95\textwidth}\footnotesize
\textit{Note.} 200 replicates per setting; Cov., coverage of nominal 95\% intervals. Largest Monte Carlo standard errors: bias 0.008, coverage 0.019.
\end{minipage}
\end{table}

\begin{table}[htbp]
\centering
\caption{Estimation of the IDE across methods, confounding strength of the lost code ($\kappa$), and sample size.}
\label{tab:core_IDE}
\small
\begin{tabular}{lrrrrrr}
\toprule
 & \multicolumn{3}{c}{$n=1,000$} & \multicolumn{3}{c}{$n=3,000$} \\
Method & Bias & RMSE & Cov. & Bias & RMSE & Cov. \\
\midrule
\multicolumn{7}{l}{\emph{$\kappa=0$}}\\
\quad Oracle $(U,R)$ & 0.003 & 0.086 & 0.975 & 0.000 & 0.059 & 0.935 \\
\quad Raw covariates $W$ & 0.019 & 0.080 & 0.940 & 0.007 & 0.055 & 0.910 \\
\quad VAE (R1), $d=4$ & 0.005 & 0.089 & 0.990 & 0.003 & 0.059 & 0.940 \\
\quad VAE (R1), $d$ by nuisance loss & 0.007 & 0.091 & 0.980 & 0.003 & 0.059 & 0.940 \\
\quad VAE (R2), fold-specific & 0.009 & 0.086 & 0.990 & 0.004 & 0.059 & 0.935 \\
\quad VAE (R3), full-sample & 0.006 & 0.087 & 0.980 & 0.004 & 0.059 & 0.935 \\
\quad VAE (R1) $+$ key covariates & 0.005 & 0.092 & 0.990 & 0.002 & 0.060 & 0.950 \\
\quad PCA (R1) & 0.003 & 0.088 & 0.980 & 0.001 & 0.059 & 0.940 \\
\quad VAE (R3), no cross-fitting & 0.007 & 0.083 & 0.975 & 0.005 & 0.058 & 0.935 \\
\multicolumn{7}{l}{\emph{$\kappa=0.5$}}\\
\quad Oracle $(U,R)$ & $-$0.006 & 0.093 & 0.960 & $-$0.001 & 0.053 & 0.950 \\
\quad Raw covariates $W$ & $-$0.004 & 0.083 & 0.935 & 0.001 & 0.051 & 0.930 \\
\quad VAE (R1), $d=4$ & $-$0.050 & 0.110 & 0.930 & $-$0.045 & 0.071 & 0.880 \\
\quad VAE (R1), $d$ by nuisance loss & $-$0.045 & 0.109 & 0.930 & $-$0.045 & 0.071 & 0.885 \\
\quad VAE (R2), fold-specific & $-$0.047 & 0.105 & 0.945 & $-$0.044 & 0.070 & 0.895 \\
\quad VAE (R3), full-sample & $-$0.049 & 0.107 & 0.940 & $-$0.044 & 0.070 & 0.885 \\
\quad VAE (R1) $+$ key covariates & $-$0.004 & 0.098 & 0.960 & 0.001 & 0.053 & 0.945 \\
\quad PCA (R1) & $-$0.051 & 0.108 & 0.940 & $-$0.047 & 0.072 & 0.875 \\
\quad VAE (R3), no cross-fitting & $-$0.051 & 0.103 & 0.915 & $-$0.043 & 0.069 & 0.885 \\
\multicolumn{7}{l}{\emph{$\kappa=1$}}\\
\quad Oracle $(U,R)$ & 0.004 & 0.094 & 0.950 & $-$0.002 & 0.048 & 0.975 \\
\quad Raw covariates $W$ & 0.003 & 0.085 & 0.925 & 0.001 & 0.046 & 0.965 \\
\quad VAE (R1), $d=4$ & $-$0.179 & 0.204 & 0.630 & $-$0.181 & 0.188 & 0.080 \\
\quad VAE (R1), $d$ by nuisance loss & $-$0.174 & 0.202 & 0.635 & $-$0.158 & 0.168 & 0.215 \\
\quad VAE (R2), fold-specific & $-$0.174 & 0.200 & 0.635 & $-$0.179 & 0.185 & 0.080 \\
\quad VAE (R3), full-sample & $-$0.176 & 0.201 & 0.625 & $-$0.179 & 0.185 & 0.080 \\
\quad VAE (R1) $+$ key covariates & 0.007 & 0.097 & 0.950 & $-$0.000 & 0.050 & 0.970 \\
\quad PCA (R1) & $-$0.180 & 0.204 & 0.645 & $-$0.183 & 0.189 & 0.075 \\
\quad VAE (R3), no cross-fitting & $-$0.173 & 0.195 & 0.565 & $-$0.177 & 0.184 & 0.085 \\
\multicolumn{7}{l}{\emph{$\kappa=2$}}\\
\quad Oracle $(U,R)$ & 0.004 & 0.107 & 0.935 & $-$0.001 & 0.055 & 0.935 \\
\quad Raw covariates $W$ & $-$0.006 & 0.095 & 0.895 & $-$0.002 & 0.053 & 0.930 \\
\quad VAE (R1), $d=4$ & $-$0.491 & 0.502 & 0.010 & $-$0.498 & 0.501 & 0.000 \\
\quad VAE (R1), $d$ by nuisance loss & $-$0.462 & 0.477 & 0.020 & $-$0.386 & 0.390 & 0.000 \\
\quad VAE (R2), fold-specific & $-$0.484 & 0.495 & 0.010 & $-$0.496 & 0.499 & 0.000 \\
\quad VAE (R3), full-sample & $-$0.487 & 0.497 & 0.015 & $-$0.495 & 0.499 & 0.000 \\
\quad VAE (R1) $+$ key covariates & 0.009 & 0.111 & 0.950 & 0.001 & 0.055 & 0.945 \\
\quad PCA (R1) & $-$0.491 & 0.502 & 0.010 & $-$0.499 & 0.503 & 0.000 \\
\quad VAE (R3), no cross-fitting & $-$0.482 & 0.491 & 0.005 & $-$0.493 & 0.496 & 0.000 \\
\bottomrule
\end{tabular}

\vspace{0.15cm}
\begin{minipage}{0.95\textwidth}\footnotesize
\textit{Note.} 200 replicates per setting; Cov., coverage of nominal 95\% intervals. Largest Monte Carlo standard errors: bias 0.008, coverage 0.035.
\end{minipage}
\end{table}

\begin{table}[htbp]
\centering
\caption{Pathway selection among 20 candidate mediators (two true). FWER: share of replicates with at least one null pathway declared significant; power: share of the two true pathways declared significant; selection bias: mean of $|\hat\theta|-|\theta|$ over selected pathways.}
\label{tab:stc}
\small
\begin{tabular}{rrlrrr}
\toprule
$\kappa$ & $n$ & Procedure & FWER & Power & Selection bias \\
\midrule
0 & 1,000 & Same-sample ranking & 52\% & 96\% & 0.044 \\
0 & 1,000 & Select-then-confirm & 0\% & 47\% & 0.030 \\
0 & 3,000 & Same-sample ranking & 56\% & 100\% & 0.024 \\
0 & 3,000 & Select-then-confirm & 0\% & 99\% & 0.006 \\
1 & 1,000 & Same-sample ranking & 100\% & 94\% & 0.358 \\
1 & 1,000 & Select-then-confirm & 21\% & 68\% & 0.329 \\
1 & 3,000 & Same-sample ranking & 100\% & 99\% & 0.340 \\
1 & 3,000 & Select-then-confirm & 34\% & 97\% & 0.324 \\
\bottomrule
\end{tabular}
\end{table}

\begin{table}[htbp]
\centering
\caption{Portable representation (lab and diagnosis encoders only). $\Delta_{\mathrm{port}}$: mean difference in the IIE between the full and portable representations in the source; ``0 in CI'': share of replicates whose interval for $\Delta_{\mathrm{port}}$ contains zero. Target: shifted population without the survey modality.}
\label{tab:port}
\small
\begin{tabular}{rrrrrrr}
\toprule
$\kappa$ & $n$ & $\Delta_{\mathrm{port}}$ & 0 in CI & Source bias & Target bias & Target cov. \\
\midrule
0 & 1,000 & $-$0.001 & 95\% & $-$0.002 & 0.003 & 0.965 \\
0 & 3,000 & 0.000 & 97\% & $-$0.002 & 0.001 & 0.970 \\
1 & 1,000 & 0.000 & 95\% & 0.185 & 0.180 & 0.300 \\
1 & 3,000 & 0.000 & 97\% & 0.182 & 0.176 & 0.020 \\
\bottomrule
\end{tabular}
\end{table}

\bibliographystyle{plainnat}
\bibliography{references}

\end{document}